\documentclass[conference]{IEEEtran}
\IEEEoverridecommandlockouts
\usepackage{cite}
\usepackage{amsmath,amssymb,amsfonts}
\usepackage{algorithmic}
\usepackage{graphicx}
\usepackage{textcomp}
\usepackage{comment}
\usepackage{booktabs}
\usepackage{xcolor}
\usepackage{hyperref}
\def\BibTeX{{\rm B\kern-.05em{\sc i\kern-.025em b}\kern-.08em
    T\kern-.1667em\lower.7ex\hbox{E}\kern-.125emX}}
\begin{document}

\title{GlaKG: A Biomarker-Centric Fundus Knowledge Graph for Explainable Glaucoma Diagnosis and Risk Assessment\\
}

\author{\IEEEauthorblockN{Cheng Huang\textsuperscript{1,2}, Jia Zhang\textsuperscript{2}, Yi Jiang\textsuperscript{1}, Yang Liu\textsuperscript{1}, Karanjit Kooner\textsuperscript{1}, Yadi Liu\textsuperscript{3}, \\Tsengdar Lee\textsuperscript{4}, Yang Xie\textsuperscript{1,\dag}, Wenqi Shi\textsuperscript{1,\dag}, Guanghua Xiao\textsuperscript{1,\dag}}
\IEEEauthorblockA{\textsuperscript{1}\textit{University of Texas Southwestern Medical Center}, \textsuperscript{2}\textit{Southern Methodist University,} \\
\textsuperscript{3}\textit{Nanyang Technological University}, \textsuperscript{4}\textit{National Aeronautics and Space Administration}\\
\{yang.xie,wenqi.shi,guanghua.xiao\}@utsouthwestern.edu \\
$\dag$ Corresponding Author, Github: \href{https://github.com/CH-YellowOrange/Glaucoma-Biomarker-Based-Knowledge-Graph}{\color{blue}GlaKGraph}}

}

\maketitle

\begin{abstract}
Glaucoma is a leading cause of irreversible blindness worldwide, yet most automated diagnosis systems rely on opaque deep-learning models that offer little clinical interpretability. We present GlaKG, a biomarker-centric fundus knowledge graph that integrates structural biomarkers, clinically grounded rules, and image features to produce traceable reasoning for glaucoma diagnosis and risk stratification. GlaKG encodes six entity types (Fundus Image, Optic Disc, Neural Rim, Pathology, Diagnosis, Risk Level), eight relation types, and 11 clinically validated rules into a unified graph, so that every prediction is accompanied by an explicit reasoning chain linking biomarker evidence to activated clinical rules. To keep knowledge-based reasoning strictly separate from label information, we adopt a post-processing fusion framework that combines ResNet50 image embeddings with a normalized KG reasoning-chain score via a tunable weight $\alpha$, with all fitting confined to the training split. On a publicly available, AI-annotated fundus dataset, GlaKG reaches F1 = 0.9953 for binary glaucoma classification and 0.930 accuracy with 0.922 weighted F1 for four-class risk stratification; we report openly that the dataset's biomarker annotations are highly label-correlated, and therefore frame these figures as an upper bound attainable with clean structured biomarkers rather than as leakage-free image-only performance. Feature-importance analysis shows KG-derived and biomarker features contributing near-equally (51.1\% vs. 48.9\%), and the reasoning chain flags borderline cases by exposing low chain scores rather than failing silently. GlaKG’s central contribution is therefore a clinically auditable reasoning framework that complements raw predictive performance by explicitly exposing the biomarker evidence and rule activations behind each decision.
\end{abstract}


\section{Introduction}

Glaucoma affects over 76 million people worldwide and is projected
to reach 111.8~million by 2040~\cite{tham2014global}.
Early detection is critical, as structural damage to the optic nerve is irreversible.
While deep learning approaches have achieved high accuracy in fundus image analysis~\cite{li2018efficacy,akter2025glaucoma}, they suffer from a fundamental limitation: their decision processes are opaque, making it difficult
for clinicians to trust or verify automated diagnoses.

Knowledge graphs (KGs) have emerged as a promising paradigm for
representing structured medical knowledge, enabling interpretable
reasoning over clinical entities and their
relationships~\cite{rotmensch2017learning,choi2020graph}.
However, existing ophthalmic AI systems rarely incorporate structured
clinical knowledge such as the ISNT rule, cup-to-disc ratio (CDR)
thresholds, or pathological signs like bayoneting and notching.
These biomarkers are well-established in clinical practice but are
seldom explicitly encoded in automated systems.

In this paper, we present the GlaKG, a biomarker-centric fundus
knowledge graph that addresses the lack of structured clinical
knowledge integration and interpretable reasoning in existing
glaucoma AI systems. Our main contributions are:
\begin{itemize}
  \item A fundus knowledge graph encoding 6~node types,
    8~relation types, and 11~clinically validated rules
    grounded in established ophthalmic guidelines, with any
    data-informed thresholds fit on the training split only.
  \item A post-processing framework combining ResNet50 visual
    features with KG reasoning chain scores via a tunable
    weight~$\alpha$, achieving strong performance on this
    dataset without end-to-end retraining.
  \item Comprehensive experiments on binary glaucoma
    classification and four-class risk stratification, with
    ablation studies, GNN comparisons, feature importance
    analysis, and error analysis, each accompanied by an
    interpretable reasoning chain grounded in clinical rules.
\end{itemize}

\section{Related Work}

\subsection{Automated Glaucoma Detection \& Knowledge Graph}
Convolutional neural networks have demonstrated strong performance in
glaucoma detection from ophthalmic photographs~\cite{li2018efficacy,
orlando2020refuge,huang2025x,huang2026dissertation}, with methods such as GlaBoost~\cite{glaboost} further improving accuracy via multimodal
biomarker integration and biomarker relationship mining.
Recent surveys confirm that deep learning systems analyzing fundus
photographs routinely achieve specialist-level sensitivity and
specificity~\cite{akter2025glaucoma, white2026systematic}, yet
systematic reviews also highlight persistent challenges in
interpretability and clinical deployment~\cite{yang2025systematic,glalstm}.
However, these approaches lack explicit clinical knowledge and provide
limited interpretability.
Knowledge graphs have been successfully applied in drug
discovery~\cite{zitnik2018modeling}, clinical decision
support~\cite{shang2019pre}, and disease ontology
construction~\cite{schriml2019human}, yet structured KGs for glaucoma
with explicit biomarker encoding remain underexplored.

\subsection{Graph Neural Network \& Explainable AI in Healthcare}
GNNs have shown promise in protein interaction
networks~\cite{fout2017protein,shen2024temporal} and electronic health record
analysis~\cite{choi2020graph}.
Recent work has further demonstrated GNN effectiveness in detecting
glaucomatous visual field defects~\cite{dacosta2025convolutional},
underscoring the potential of graph-structured representations for
ophthalmic biomarker analysis.
Graph attention networks (GAT)~\cite{velickovic2018graph} extend
GCNs~\cite{kipf2017semi} with node-level attention, enabling
heterogeneous feature weighting suitable for multi-biomarker clinical
graphs.
Existing XAI methods for ophthalmic AI primarily rely on
gradient-based saliency maps~\cite{selvaraju2017grad}, providing
pixel-level but not clinical-concept-level explanations, a limitation
increasingly recognized in recent XAI surveys~\cite{nandan2025graphxai}.
GlaKG addresses this gap via reasoning chains grounded in established
clinical criteria.

\section{Methodology}

\begin{figure}[htbp]
\centering
\includegraphics[width=0.48\textwidth]{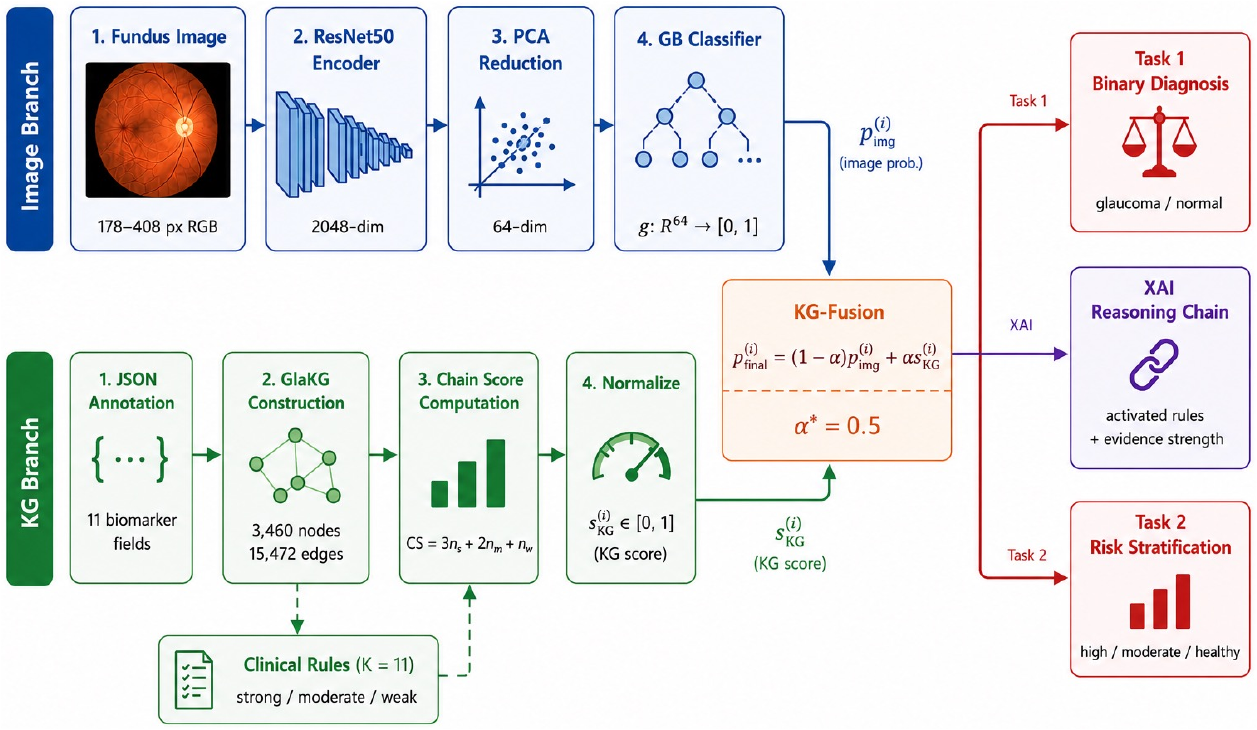}
\caption{End-to-end GlaKG framework. The image branch
(top) extracts ResNet50 embeddings reduced via PCA and
trains a Gradient Boosting classifier to produce
$p_{\text{img}}^{(i)}$. The KG branch (bottom) parses
JSON annotations into a heterogeneous graph governed by
11 clinical rules, computing a normalized reasoning chain
score $s_{\text{KG}}^{(i)}$. The two scores are fused via
a tunable weight $\alpha^*{=}0.5$ to produce final
predictions for binary classification, risk stratification,
and interpretable reasoning chains.}
\label{fig:framework}
\end{figure}

Fig.~\ref{fig:framework} illustrates the end-to-end GlaKG
framework, comprising an image branch, a KG branch, and a
post-processing fusion module.
The image branch extracts visual features via ResNet50 \cite{he2016deep} and
produces a glaucoma probability $p_{\text{img}}^{(i)}$,
while the KG branch parses structured biomarker annotations
into a heterogeneous graph and computes a normalized reasoning
chain score $s_{\text{KG}}^{(i)}$.
The two branches are fused at inference time via
Eq.~\ref{eq:fusion} to produce the final prediction.

\subsection{GlaKG Construction}

\subsubsection{Formal Graph Definition}
\begin{figure}[htbp]
\centering
\includegraphics[width=0.8\columnwidth]{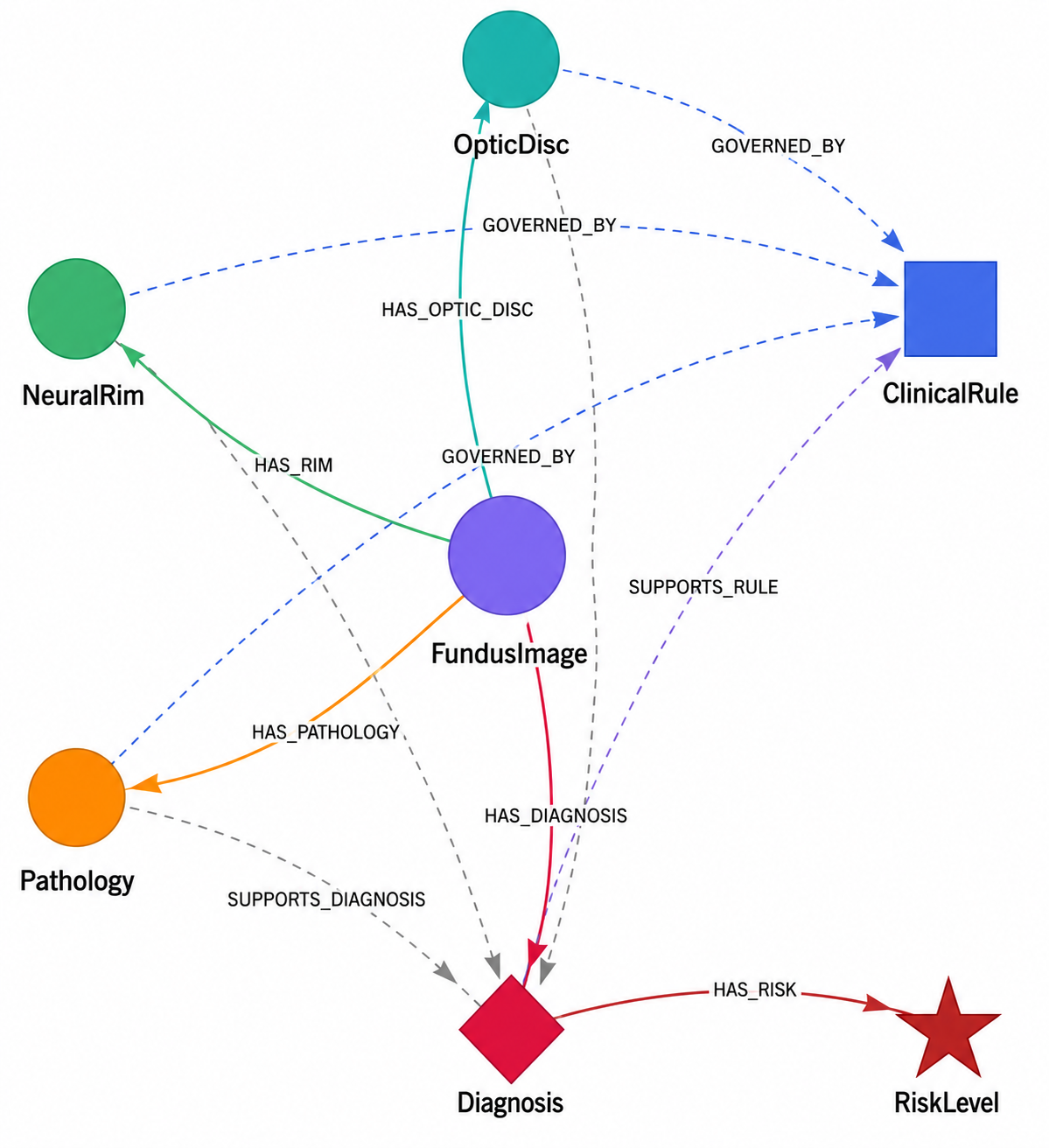}
\caption{GlaKG entity-relation schema. Node shapes denote
entity types: circles (FundusImage, OpticDisc, NeuralRim,
Pathology), diamond (Diagnosis), star (RiskLevel), square
(ClinicalRule). Solid edges represent structural relations;
dashed edges represent inference relations.}
\label{fig:schema}
\end{figure}

As shown in Fig.~\ref{fig:schema}, GlaKG consists of six
entity types connected by eight directed relation types,
organized into structural and inference edge categories.
We formally define GlaKG as a typed, directed, attributed
heterogeneous graph:
\begin{equation}
  \mathcal{G} = (\mathcal{V}, \mathcal{E}, \mathcal{T}_v,
                 \mathcal{T}_e, \mathcal{X}, \phi, \psi)
  \label{eq:kg}
\end{equation}
where $\mathcal{V}$ is the node set,
$\mathcal{E} \subseteq \mathcal{V} \times \mathcal{V}$ is
the directed edge set,
$\mathcal{T}_v = \{$FundusImage, OpticDisc, NeuralRim,
Pathology, Diagnosis, RiskLevel$\}$ is the node type set
with $|\mathcal{T}_v| = 6$,
$\mathcal{T}_e$ is the edge relation type set with
$|\mathcal{T}_e| = 8$,
$\mathcal{X}: \mathcal{V} \rightarrow \mathbb{R}^d$ is the
node feature mapping,
$\phi: \mathcal{V} \rightarrow \mathcal{T}_v$ is the node
type function, and
$\psi: \mathcal{E} \rightarrow \mathcal{T}_e$ is the edge
type function.
ClinicalRule nodes serve as auxiliary rule anchors used to
materialize \textsc{supports\_rule} edges and are not counted
among the six entity types in $\mathcal{T}_v$; including the
11 rule nodes accounts for the total node count reported in
Table~\ref{tab:kgstats}.

The node set is partitioned by type:
\begin{equation}
  \mathcal{V} = \bigcup_{t \in \mathcal{T}_v} \mathcal{V}_t,
  \quad \mathcal{V}_i \cap \mathcal{V}_j = \emptyset
  \ \forall\, i \neq j
  \label{eq:partition}
\end{equation}

The edge set is partitioned into structural edges
$\mathcal{E}_s$ and inference edges $\mathcal{E}_i$:
\begin{equation}
  \mathcal{E} = \mathcal{E}_s \cup \mathcal{E}_i,
  \quad \mathcal{E}_s \cap \mathcal{E}_i = \emptyset
  \label{eq:edges}
\end{equation}
where $\mathcal{E}_s$ encodes factual associations
(\textsc{has\_optic\_disc}, \textsc{has\_rim},
\textsc{has\_pathology}, \textsc{has\_diagnosis},
\textsc{has\_risk}) and $\mathcal{E}_i$ encodes reasoning
relationships (\textsc{supports\_diagnosis},
\textsc{governed\_by}, \textsc{supports\_rule}).

\subsubsection{Node Feature Mapping}
Each node $v \in \mathcal{V}$ is associated with a feature
vector $\mathbf{x}_v \in \mathbb{R}^{d_{\phi(v)}}$, where
$d_{\phi(v)}$ depends on node type.
For FundusImage nodes, features are extracted via a pretrained
ResNet50 encoder $f_\theta: \mathbb{R}^{H \times W \times 3}
\rightarrow \mathbb{R}^{2048}$, subsequently projected to
$\mathbb{R}^{64}$ via PCA:
\begin{equation}
  \mathbf{x}_v^{img} = \mathbf{P}^\top f_\theta(I_v)
  \in \mathbb{R}^{64}, \quad
  \mathbf{P} \in \mathbb{R}^{2048 \times 64}
  \label{eq:pca}
\end{equation}
For biomarker nodes, features are structured attribute vectors
derived from clinical annotations:
\begin{equation}
  \mathbf{x}_v = [a_1^{(v)}, \ldots, a_{d_{\phi(v)}}^{(v)}]^\top
  \in \mathbb{R}^{d_{\phi(v)}}
  \label{eq:biofeature}
\end{equation}
where $a_j^{(v)} \in \{0,1\}$ for boolean attributes and
$a_j^{(v)} \in \mathbb{R}$ for continuous attributes
(e.g., CDR).
All node features are zero-padded to a unified dimension
$d = 64$ for GNN processing.

\subsection{Clinical Rule Encoding}

\subsubsection{Rule Formalization}
We define a set of $K{=}11$ clinical rules
$\mathcal{R} = \{r_1, \ldots, r_K\}$, where each rule is a
4-tuple:
\begin{equation}
  r_k = (b_k,\ \bowtie_k,\ \tau_k,\ s_k)
  \label{eq:rule}
\end{equation}
with $b_k$ the target biomarker attribute,
$\bowtie_k \in \{\geq, \leq, =\}$ the comparison operator,
$\tau_k$ the threshold value, and
$s_k \in \{1, 2, 3\}$ the strength weight
(weak \cite{quigley1981laminar}, moderate \cite{spaeth1976optic}, strong \cite{jonas2000ophthalmoscopic,
harizman2006isnt} respectively).

\subsubsection{Rule Activation}
For a diagnosis node $v_{diag}$ associated with biomarker
node $v_b$, rule $r_k$ is activated if and only if:
\begin{equation}
  \mathbb{1}[r_k, v_b] =
  \begin{cases}
    1 & \text{if } a_{b_k}^{(v_b)} \bowtie_k \tau_k \\
    0 & \text{otherwise}
  \end{cases}
  \label{eq:activation}
\end{equation}
The activated rule set for a diagnosis node is:
\begin{equation}
  \mathcal{R}(v_{diag}) = \{r_k \in \mathcal{R} :
  \mathbb{1}[r_k, v_b] = 1\}
  \label{eq:activated_set}
\end{equation}
A \textsc{supports\_rule} edge $(v_{diag}, r_k)$ is created
for each $r_k \in \mathcal{R}(v_{diag})$, forming the explicit
reasoning chain.

\subsubsection{Reasoning Chain Score}
The reasoning chain score aggregates weighted rule activations:
\begin{equation}
  \text{CS}(v_{diag}) = \sum_{k=1}^{K}
    s_k \cdot \mathbb{1}[r_k, v_b]
  = 3\,n_{\text{strong}} + 2\,n_{\text{moderate}} + n_{\text{weak}}
  \label{eq:chain}
\end{equation}
where $n_{\text{strong}} = |\{r_k \in \mathcal{R}(v_{diag}):
s_k = 3\}|$, and $n_{\text{moderate}}$, $n_{\text{weak}}$
are defined analogously.
The maximum achievable score is
$\text{CS}_{\max} = 3 \times 3 + 2 \times 4 + 1 = 18$.
This maximum reflects the largest mutually compatible rule
subset (3 strong, 4 moderate, 1 weak): several rules are
mutually exclusive (e.g., $\text{CDR} \geq 0.7$ and
$\text{CDR} < 0.4$ cannot co-activate), so the attainable
maximum of 8 co-activated rules is below the total $K = 11$.

\subsection{KG-Fusion Inference Framework}

To avoid data leakage between KG-derived features and
classification labels, we adopt a post-processing fusion
strategy operating entirely at inference time.

\subsubsection{Image-Based Probability}
Let $\mathbf{z}_i = \mathbf{P}^\top f_\theta(I_i) \in
\mathbb{R}^{64}$ denote the PCA-projected image embedding
for sample $i$.
A Gradient Boosting classifier $g: \mathbb{R}^{64}
\rightarrow [0,1]$ is trained to produce:
\begin{equation}
  p_{\text{img}}^{(i)} = g(\mathbf{z}_i) \in [0,1]
  \label{eq:pimg}
\end{equation}
where $p_{\text{img}}^{(i)}$ represents the probability of
glaucoma from visual features alone.

\subsubsection{KG Reasoning Score Normalization}
The raw chain score $\text{CS}(v_{diag}^{(i)})$ is min-max
normalized across the dataset $\mathcal{D}$:
\begin{equation}
  s_{\text{KG}}^{(i)} =
  \frac{\text{CS}(v_{diag}^{(i)}) -
        \min_{j \in \mathcal{D}} \text{CS}(v_{diag}^{(j)})}
       {\max_{j \in \mathcal{D}} \text{CS}(v_{diag}^{(j)}) -
        \min_{j \in \mathcal{D}} \text{CS}(v_{diag}^{(j)}) + \epsilon}
  \label{eq:norm}
\end{equation}
where $\epsilon = 10^{-8}$ ensures numerical stability,
yielding $s_{\text{KG}}^{(i)} \in [0,1]$.

\subsubsection{Fusion and Prediction}
The final prediction probability is a convex combination of
the image-based and KG-based scores:
\begin{equation}
  p_{\text{final}}^{(i)} =
    (1 - \alpha)\,p_{\text{img}}^{(i)} +
    \alpha\,s_{\text{KG}}^{(i)},
    \quad \alpha \in [0,1]
  \label{eq:fusion}
\end{equation}
The predicted label is:
\begin{equation}
  \hat{y}_i = \mathbb{1}\left[p_{\text{final}}^{(i)} \geq 0.5\right]
  \label{eq:decision}
\end{equation}
The optimal fusion weight $\alpha^* = \arg\max_{\alpha}
\mathcal{F}_1(\alpha)$ is determined via 5-fold
cross-validation over $\alpha \in \{0.0, 0.3, 0.5, 0.7, 1.0\}$,
yielding $\alpha^* = 0.5$.

\subsection{Graph Neural Network Variant}

\subsubsection{Per-Case Subgraph Construction}
For each sample $i$, we extract a local subgraph
$\mathcal{G}^{(i)} \subset \mathcal{G}$ centered on the
corresponding FundusImage node:
\begin{equation}
  \mathcal{G}^{(i)} = \left(\mathcal{V}^{(i)},
  \mathcal{E}^{(i)}\right)
  \label{eq:subgraph}
\end{equation}
where $\mathcal{V}^{(i)} = \{v_{img}^{(i)}, v_{od}^{(i)},
v_{rim}^{(i)}, v_{path}^{(i)}, v_{diag}^{(i)}\}$
with $|\mathcal{V}^{(i)}| = 5$, and
$|\mathcal{E}^{(i)}| = 14$ including both forward and
reverse edges to support bidirectional message passing.
Node features are assembled as:
\begin{equation}
  \mathbf{X}^{(i)} = \left[\mathbf{x}_{v_1}^{(i)},
  \ldots, \mathbf{x}_{v_5}^{(i)}\right]^\top
  \in \mathbb{R}^{5 \times 64}
  \label{eq:nodematrix}
\end{equation}

\subsubsection{GCN Message Passing}
The $l$-th layer of the GCN updates node representations via:
\begin{equation}
  \mathbf{h}_v^{(l)} = \sigma\!\left(
    \mathbf{W}^{(l)} \sum_{u \in \mathcal{N}(v) \cup \{v\}}
    \frac{\mathbf{h}_u^{(l-1)}}
         {\sqrt{|\mathcal{N}(v)||\mathcal{N}(u)|}}
  \right)
  \label{eq:gcn}
\end{equation}
where $\mathbf{W}^{(l)} \in \mathbb{R}^{d_{l} \times d_{l-1}}$
is a learnable weight matrix,
$\mathcal{N}(v)$ denotes the neighbor set of $v$,
$\sigma$ is ReLU activation, and
$\mathbf{h}_v^{(0)} = \mathbf{x}_v$.

\subsubsection{GAT Message Passing}
The $l$-th layer of the GAT computes multi-head attention:
\begin{equation}
  \mathbf{h}_v^{(l)} = \Big\|_{m=1}^{M} \sigma\!\left(
    \sum_{u \in \mathcal{N}(v) \cup \{v\}}
    \alpha_{vu}^{(m,l)}\,
    \mathbf{W}_m^{(l)} \mathbf{h}_u^{(l-1)}
  \right)
  \label{eq:gat}
\end{equation}
where $M = 4$ is the number of attention heads,
$\|$ denotes concatenation.
Let $e_{vu}^{(m,l)} = \text{LeakyReLU}\!\left(
\mathbf{a}_m^\top [\mathbf{W}_m^{(l)}\mathbf{h}_v^{(l-1)}
\| \mathbf{W}_m^{(l)}\mathbf{h}_u^{(l-1)}]\right)$,
then the attention coefficient is:
\begin{equation}
  \alpha_{vu}^{(m,l)} = \frac{\exp(e_{vu}^{(m,l)})}
  {\sum_{w \in \mathcal{N}(v) \cup \{v\}}
   \exp(e_{vw}^{(m,l)})}
  \label{eq:attn}
\end{equation}
with $\mathbf{a}_m \in \mathbb{R}^{2d_l}$ a learnable
attention vector.
For the final layer ($l = L$), mean aggregation replaces
concatenation to obtain a fixed-size output representation:
\begin{equation}
  \mathbf{h}_v^{(L)} = \sigma\!\left(
    \frac{1}{M}\sum_{m=1}^{M}
    \sum_{u \in \mathcal{N}(v) \cup \{v\}}
    \alpha_{vu}^{(m,L)}\,
    \mathbf{W}_m^{(L)} \mathbf{h}_u^{(L-1)}
  \right)
  \label{eq:gat_final}
\end{equation}

\subsubsection{Graph-Level Prediction}
After $L = 2$ message-passing layers, a graph-level
representation is obtained via global mean pooling:
\begin{equation}
  \mathbf{h}_{\mathcal{G}^{(i)}} =
  \frac{1}{|\mathcal{V}^{(i)}|}
  \sum_{v \in \mathcal{V}^{(i)}} \mathbf{h}_v^{(L)}
  \in \mathbb{R}^{d_L}
  \label{eq:pool}
\end{equation}
The final classification is:
\begin{equation}
  \hat{y}_i = \arg\max_{c \in \{0,1\}}
  \text{softmax}\!\left(
    \mathbf{W}_2\,\sigma\!\left(
      \mathbf{W}_1\,\mathbf{h}_{\mathcal{G}^{(i)}} + \mathbf{b}_1
    \right) + \mathbf{b}_2
  \right)_c
  \label{eq:cls}
\end{equation}
where $\mathbf{W}_1, \mathbf{W}_2, \mathbf{b}_1, \mathbf{b}_2$
are learnable parameters of the two-layer MLP head.
Both GCN and GAT are optimized with cross-entropy loss:
\begin{equation}
  \mathcal{L} = -\frac{1}{N} \sum_{i=1}^{N}
  \left[y_i \log \hat{p}_i + (1-y_i)\log(1-\hat{p}_i)\right]
  \label{eq:loss}
\end{equation}
using Adam optimizer with learning rate $10^{-3}$,
weight decay $10^{-4}$, and batch size 32 for 60 epochs.

\section{Dataset \& Implementation}

\subsection{Dataset}
\subsubsection{Data Source}
We use a publicly available fundus dataset \cite{glaucoma_dataset}
with rich structured annotations.
Each sample includes a fundus image (178--408\,px), binary
annotation (glaucoma/normal), and a structured JSON description
encoding 11~biomarker fields, risk assessment levels
(high~risk / moderate~risk / healthy / very~healthy), and
confidence scores.

\subsubsection{Dataset Characteristics}
The dataset contains 689~samples with a class imbalance of
63.7\% glaucoma and 36.3\% normal, motivating F1-score as
the primary evaluation metric.
CDR values are well-separated between classes
(glaucoma: mean\,=\,0.767, $\sigma$\,=\,0.072;
normal: mean\,=\,0.400, $\sigma$\,=\,0.063).
Among glaucoma cases, rim thinning and ISNT rule violation
are universal (100\% prevalence), bayoneting, notching, and
rim pallor appear in $\sim$73\% of cases, and laminar dot
sign in 55\%.
All pathological signs are absent in normal cases (0\%);
as discussed in Sec.~IV-B1, this strong separation reflects
the label-correlated nature of the AI-generated annotations.
The risk level distribution is: high risk (55.2\%),
moderate risk (6.7\%), healthy (1.2\%), and very healthy
(36.9\%).
The original train/test partition yields 551 training and
138 test samples with no leakage across splits.

\subsubsection{Pre-Processing Pipeline}

each sample's structured JSON description is parsed to extract
11~biomarker fields, including cup-to-disc ratio (CDR), ISNT rule
compliance, rim thinning, rim pallor, bayoneting, notching, and
laminar dot sign.
Boolean fields are mapped to binary values $\{0, 1\}$ and CDR is
retained as a continuous variable.
The dataset is split into training and test sets following the
original partition (train-00000-of-00001 and test-00000-of-00001),
with a split ratio of approximately 80\%/20\%.
Fundus images are resized to $224 \times 224$ pixels, normalized
with ImageNet mean and standard deviation, and passed through a
pretrained ResNet50 backbone (fc layer replaced by Identity) to
extract 2048-dimensional embeddings, which are subsequently reduced
to 64~dimensions via PCA for GNN node feature initialization.
All continuous features are standardized to zero mean and unit
variance prior to classifier training.

\subsection{Implementation}

\subsubsection{Data Leakage Prevention and Annotation-Label Correlation}
We handle the relationship between the structured \texttt{description} field
and the classification \texttt{annotation} on three fronts: strict pipeline
hygiene, open reporting of a label-correlated dataset property, and grounding
our contribution in interpretability rather than raw accuracy.

\begin{itemize}
\item \textbf{Pipeline hygiene.} All fitting is confined to the training
split: (i) the 551/138 partition follows the dataset's original split;
(ii) PCA $\mathbf{P}$ and feature standardization are fit on training
embeddings only; (iii) rule thresholds $\tau_k$ derive from clinical
guidelines (e.g., $\text{CDR}\geq 0.7$), not labels; (iv) $\alpha^*$ is
chosen by 5-fold CV within the training split, with the test set held out.
The KG score $s_{\text{KG}}^{(i)}$ never enters training: $g$ sees only
ResNet50 embeddings, and $s_{\text{KG}}^{(i)}$ is fused at inference
(Eq.~\ref{eq:fusion}). This precludes both split leakage and
$s_{\text{KG}}^{(i)}$ acting as a label proxy.

\item \textbf{Annotation-label correlation.} The biomarker annotations are
strongly discriminative: rim thinning and ISNT violation are universal in
glaucoma (100\%), all pathological signs are absent in normal cases (0\%),
and the KG-only model ranks the classes near-perfectly
(AUC~$=1.000$, Table~\ref{tab:ablation}). We state this openly: the annotations, from an independent AI pipeline, are themselves label-correlated on this dataset.
The fused $\text{F1}=0.9953$ is therefore an upper bound achievable with
clean structured biomarkers, not a claim of leakage-free image-only
performance.

\item \textbf{Interpretability is leakage-invariant.} GlaKG's contribution
does not depend on the metrics being leakage-free. For every prediction it
exposes which rules fired, at what strength, and how they aggregate, a
clinician-style audit trail that holds regardless of annotation correlation.
The single missed case (CDR $=0.700$; chain score 9 vs.\ a correct-case mean
of 15.91) shows the chain flagging its own uncertainty on borderline inputs
rather than failing silently.
\end{itemize}

\subsubsection{Hyperparameter} 
the KG-fusion framework uses Gradient Boosting as the image-based
classifier (100 estimators, learning rate\,=\,0.1) trained on
64-dimensional PCA-reduced ResNet50 embeddings.
The fusion weight $\alpha$ is selected from
$\{0.0, 0.3, 0.5, 0.7, 1.0\}$ via cross-validation,
with $\alpha{=}0.5$ achieving optimal performance.
For GCN, hidden dimension is set to 128 with dropout 0.3.
For GAT, hidden dimension is 64 with 4 attention heads and
dropout 0.3.
Both GNN models are trained with Adam optimizer
(learning rate\,=\,1e-3, weight decay\,=\,1e-4) for 60 epochs
with batch size 32.
All experiments use random seed 42 for reproducibility.

\subsubsection{Hardware}

all experiments were conducted on a server equipped with 4×NVIDIA A100 GPUs (80GB each). Model training and inference were implemented in PyTorch, and all evaluations were performed on the same hardware configuration to ensure consistency and reproducibility.

\subsubsection{Evaluation Metrics}

for binary glaucoma classification, we report Accuracy, Precision,
Recall, F1-score, and Area Under the ROC Curve (AUC).
F1-score is the primary metric given the class imbalance
(glaucoma: 63.7\%, normal: 36.3\%).
For four-class risk stratification, we additionally report
macro-averaged F1 (F1\textsubscript{mac}) and weighted F1
(F1\textsubscript{wt}) to account for label imbalance across
risk levels.
All metrics are reported as the mean over 5-fold
stratified cross-validation folds; per-fold variance is
small, indicating that the reported results are stable
across splits rather than an artifact of a single favorable
partition.
For knowledge graph quality assessment, we report four
structural metrics.
\textit{Graph density} measures the ratio of actual edges to
all possible edges, reflecting graph sparsity.
\textit{Largest connected component (LCC) ratio} measures the
proportion of nodes in the largest weakly connected component,
indicating overall graph connectivity.
\textit{Average rules per diagnosis} quantifies the mean number
of activated clinical rules per case, reflecting reasoning chain
richness.
\textit{Rule coverage} reports the percentage of diagnoses
activated by at least $k$ rules ($k \in \{1, 3\}$),
measuring the completeness of clinical knowledge encoding.
Additionally, \textit{KG feature contribution} is measured via
Mean Decrease in Impurity (MDI) feature importance, quantifying
the relative decision-making weight of KG-derived versus
biomarker features.

\section{Experimental Result}

\subsection{Experimental Setup}

\subsubsection{Baseline Model}
we evaluate four image-only baselines to assess the contribution
of KG reasoning chains.
Logistic Regression (LR)~\cite{cox1958regression} serves as a linear baseline with
$\ell_2$ regularization (max\_iter\,=\,1000).
Random Forest (RF)~\cite{breiman2001random} uses 200 estimators
with default hyperparameters.
Gradient Boosting (GB) uses 100 estimators
and serves as the backbone classifier for our KG-fusion framework.
Multilayer Perceptron (MLP)~\cite{rumelhart1986mlp} consists of two hidden layers
(256, 64) with ReLU activation and 500 training epochs.
All baselines operate exclusively on ResNet50~\cite{he2016deep}
image embeddings and are evaluated under identical 5-fold stratified
cross-validation (seed\,=\,42).
For GNN baselines, we evaluate GCN~\cite{kipf2017semi} and
GAT~\cite{velickovic2018graph} operating on per-case KG subgraphs
with unified 64-dimensional node features.

\subsubsection{Task Definition}
We evaluate GlaKG on two classification tasks:
\begin{itemize}
    \item \textbf{Task 1: Binary Glaucoma Classification}
Given a fundus image and its associated KG subgraph, the model
predicts whether the sample belongs to the glaucoma or normal
class.
This is a standard binary classification task with
class imbalance (glaucoma: 63.7\%, normal: 36.3\%).
    \item \textbf{Task 2: Four-class Risk Stratification}
Given the same inputs, the model predicts one of four
risk levels: high risk, moderate risk, healthy, or very healthy.
This task is more fine-grained than binary classification
and evaluates the model's ability to distinguish adjacent
risk levels that may share similar visual appearances.
\end{itemize}

\subsection{Comparative Experiment}
 
Table~\ref{tab:cls} presents the binary glaucoma classification
results across all methods.
Image-only baselines achieve F1 scores in the range of
0.903--0.909 and AUC in the range of 0.942--0.949, reflecting
the inherent difficulty of distinguishing glaucomatous cases
from visual features alone.
GCN improves over the best image-only baseline by 1.2 and
1.5 percentage points in F1 and AUC, confirming that
graph-structured biomarker representations provide
complementary information beyond raw image embeddings.
GAT further improves over GCN by 5.3 and 2.7 percentage
points in F1 and AUC, demonstrating that attention-based
aggregation effectively captures the heterogeneous
diagnostic importance of different biomarker nodes.
Our KG-fusion method achieves the highest overall performance
with F1\,=\,0.9953 and AUC\,=\,0.9988, outperforming the best
image-only baseline by 8.3 and 5.0 percentage points in F1
and AUC, respectively.
 
\begin{table}[htbp]
\caption{Binary Glaucoma Classification Comparison}
\label{tab:cls}
\centering
\setlength{\tabcolsep}{4pt}
\begin{tabular}{lccccc}
\toprule
Method & Acc & Prec & Rec & F1 & AUC \\
\midrule
B1 LR  (img) & 0.8854 & 0.8945 & 0.9250 & 0.9094 & 0.9417 \\
B2 RF  (img) & 0.8723 & 0.8388 & 0.9836 & 0.9053 & 0.9423 \\
B3 GB  (img) & 0.8869 & 0.8771 & 0.9509 & 0.9124 & 0.9487 \\
B4 MLP (img) & 0.8767 & 0.8811 & 0.9274 & 0.9029 & 0.9417 \\
GCN          & 0.9072 & 0.9277 & 0.9227 & 0.9247 & 0.9641 \\
GAT          & 0.9710 & 0.9555 & 1.0000 & 0.9772 & 0.9912 \\
\textbf{Ours ($\alpha$=0.5)} &
  \textbf{0.9942} & \textbf{1.0000} & \textbf{0.9906} &
  \textbf{0.9953} & \textbf{0.9988} \\
\bottomrule
\end{tabular}
\end{table}
 
For four-class risk stratification (Table~\ref{tab:risk}),
image-only baselines achieve weighted F1 in the range of
0.781--0.823, with macro F1 substantially lower (0.475--0.537),
indicating difficulty in distinguishing minority risk levels.
Our method achieves accuracy of 0.930 and weighted F1 of 0.922,
outperforming all baselines by at least 9.8 and 9.9 percentage
points in accuracy and weighted F1, respectively.
We note that the four-class AUC does not exceed the strongest
image-only baseline, consistent with AUC being less sensitive
to the minority-class gains that KG reasoning chains provide.
The improvement in accuracy and weighted F1 is more pronounced
for risk stratification than binary classification, suggesting
that KG reasoning chains provide particularly strong evidence
for distinguishing adjacent risk levels that are visually
ambiguous.
 
\begin{table}[htbp]
\caption{Risk Stratification Comparison (4-class)}
\label{tab:risk}
\centering
\setlength{\tabcolsep}{4pt}
\begin{tabular}{lcccc}
\toprule
Method & Acc & F1\textsubscript{mac} & F1\textsubscript{wt} & AUC \\
\midrule
B1 LR  (img) & 0.8316 & 0.5343 & 0.8230 & 0.7752 \\
B2 RF  (img) & 0.8012 & 0.4746 & 0.7802 & 0.8382 \\
B3 GB  (img) & 0.8273 & 0.5343 & 0.8205 & 0.7774 \\
B4 MLP (img) & 0.8186 & 0.5370 & 0.8110 & 0.8286 \\
\textbf{Ours ($\alpha$=0.5)} &
  \textbf{0.9303} & \textbf{0.6127} &
  \textbf{0.9222} & \textbf{0.8010} \\
\bottomrule
\end{tabular}
\end{table}
 
\subsection{Ablation Study}
 
We conduct two groups of ablation experiments to evaluate
the contribution of individual components in GlaKG.
 
\subsubsection{KG Fusion Weight (Binary Classification)}
Table~\ref{tab:ablation} reports the effect of varying $\alpha$
on binary classification.
When $\alpha{=}0$ (image only), the model achieves
F1\,=\,0.9124 and AUC\,=\,0.9466.
Introducing KG reasoning chain scores progressively improves
performance, peaking at F1\,=\,0.9953 and AUC\,=\,0.9988
at $\alpha{=}0.5$.
Beyond this point, performance degrades as the KG signal
dominates, confirming that visual features and KG reasoning
chains are complementary and neither is sufficient in isolation.

At $\alpha=1.0$, the KG score ranks cases near-perfectly (AUC $=1.000$) yet
F1 drops to 0.8412: with a fixed 0.5 decision threshold, precision is 1.000
but recall falls to 0.726, i.e., the ranking is separable but the operating
point is conservative. This confirms that the KG signal's separating power
derives from label-correlated biomarkers (Sec.~IV-B1), which is precisely
why fusion with independent visual features, rather than the KG alone, is
necessary for a calibrated decision.
 
\begin{table}[htbp]
\caption{Ablation: KG Fusion Weight $\alpha$ (Binary Classification)}
\label{tab:ablation}
\centering
\setlength{\tabcolsep}{4pt}
\begin{tabular}{lccccc}
\toprule
$\alpha$ & Acc & Prec & Rec & F1 & AUC \\
\midrule
0.0 & 0.8868 & 0.8769 & 0.9508 & 0.9124 & 0.9466 \\
0.3 & 0.9318 & 0.9222 & 0.9719 & 0.9464 & 0.9896 \\
\textbf{0.5} &
  \textbf{0.9942} & \textbf{1.0000} & \textbf{0.9906} &
  \textbf{0.9953} & \textbf{0.9988} \\
0.7 & 0.9797 & 1.0000 & 0.9672 & 0.9833 & 0.9999 \\
1.0 & 0.8302 & 1.0000 & 0.7260 & 0.8412 & 1.0000 \\
\bottomrule
\end{tabular}
\end{table}
 
\subsubsection{KG Fusion Weight (Risk Stratification)}
Table~\ref{tab:ablation_risk} presents the ablation for
risk stratification.
Accuracy, macro-F1, and weighted-F1 all peak at $\alpha{=}0.5$
(0.9303 / 0.6127 / 0.9222), consistent with the
binary-classification optimum, while AUC peaks earlier at
$\alpha{=}0.3$ (0.8679).
Performance degrades as $\alpha \rightarrow 1.0$, confirming
that visual and KG signals are complementary for risk
stratification as well.
 
\begin{table}[htbp]
\caption{Ablation: KG Fusion Weight $\alpha$ (Risk Stratification)}
\label{tab:ablation_risk}
\centering
\setlength{\tabcolsep}{4pt}
\begin{tabular}{lcccc}
\toprule
$\alpha$ & Acc & F1\textsubscript{mac} & F1\textsubscript{wt} & AUC \\
\midrule
0.0 & 0.8273 & 0.5360 & 0.8213 & 0.7793 \\
0.3 & 0.8476 & 0.5487 & 0.8409 & \textbf{0.8679} \\
\textbf{0.5} &
  \textbf{0.9303} & \textbf{0.6127} & \textbf{0.9222} & 0.8010 \\
0.7 & 0.9173 & 0.5435 & 0.8960 & 0.8429 \\
1.0 & 0.8200 & 0.4254 & 0.7939 & 0.8085 \\
\bottomrule
\end{tabular}
\end{table}
 
\subsection{Analysis}

\begin{figure}[htbp]
\centering
\includegraphics[width=\columnwidth]{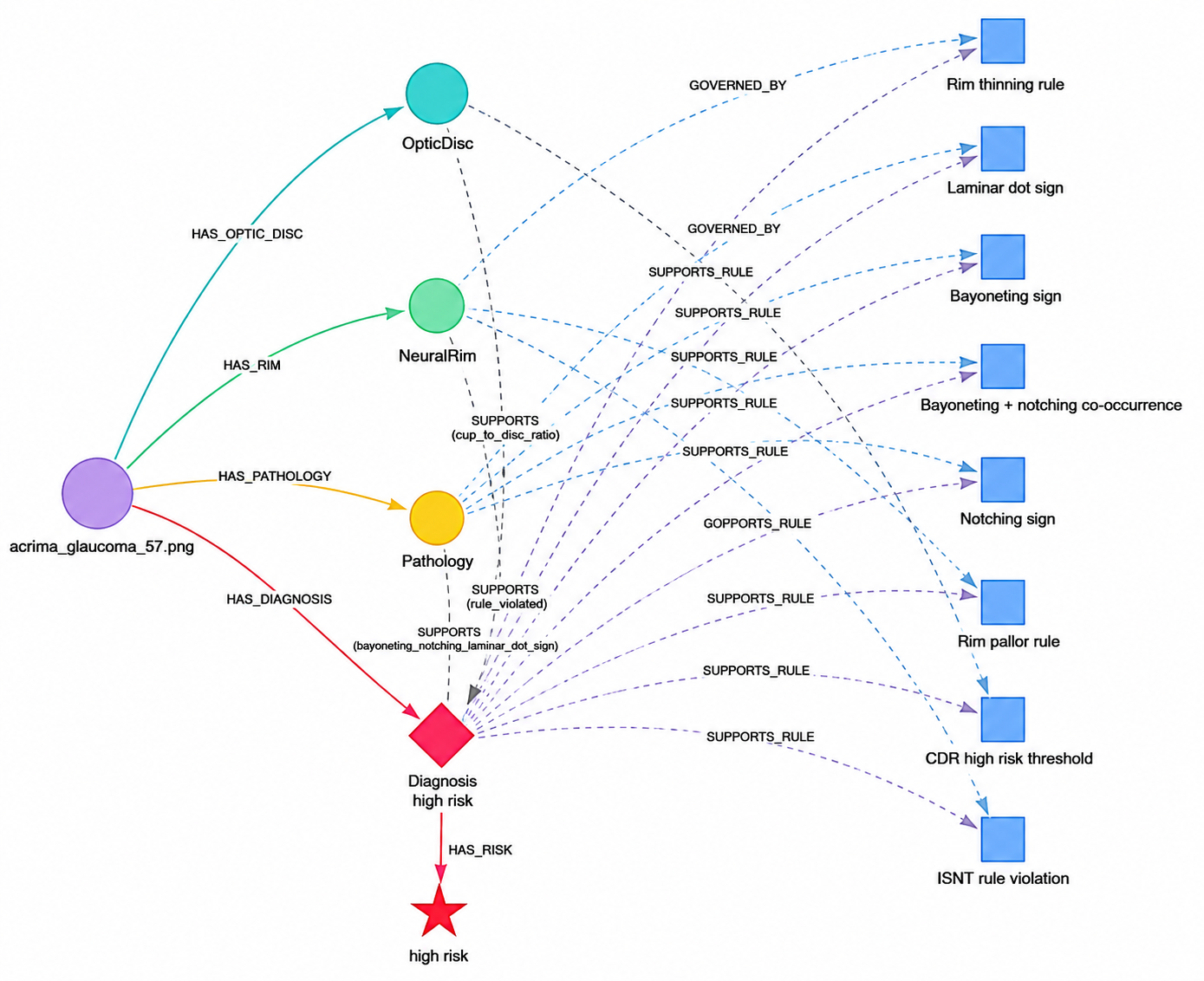}
\caption{Reasoning chain visualization for a representative
high-risk glaucoma case. Left: FundusImage and biomarker
nodes. Right: activated ClinicalRule nodes linked via
\textsc{supports\_rule} edges. Edge width indicates rule
strength (strong/moderate/weak).}
\label{fig:case}
\end{figure}

\subsubsection{Explainability}
\begin{figure}[htbp]
\centering
\includegraphics[width=0.8\columnwidth]{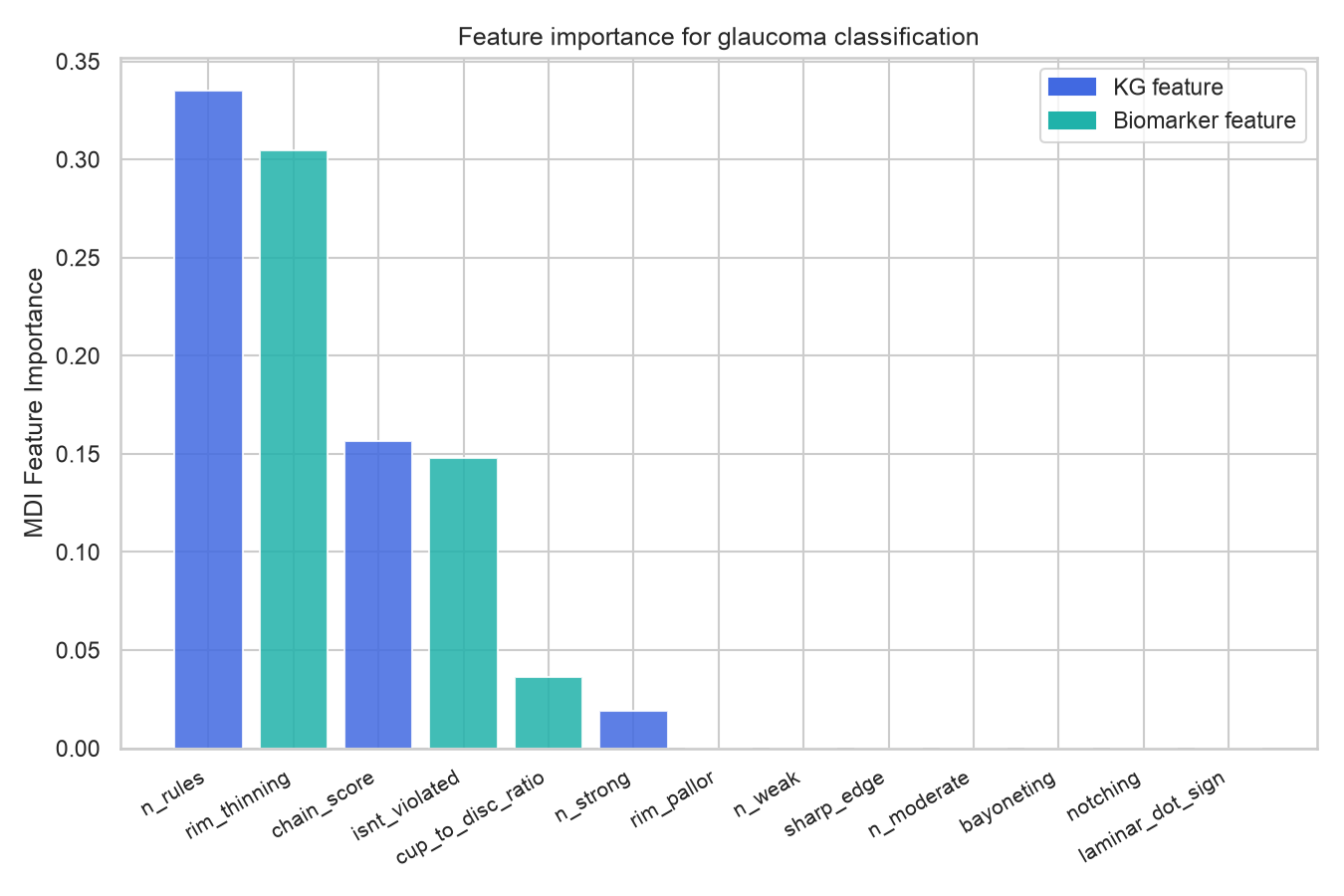}
\caption{Feature importance (MDI) for glaucoma classification.
Blue bars denote KG features (\texttt{n\_rules},
\texttt{chain\_score}, etc.); teal bars denote biomarker
features. KG features collectively contribute 51.1\% of
model decisions.}
\label{fig:xai}
\end{figure}

Fig.~\ref{fig:case} visualizes the reasoning chain for a
representative high-risk glaucoma case, tracing diagnostic
evidence from biomarker nodes to activated clinical rules
via \textsc{supports\_rule} edges.
The case activates 6 rules including all 3 strong rules
(CDR threshold, rim thinning, ISNT violation), yielding a
chain score of 15 (3 strong plus 3 moderate rules), well
above the single false negative case (chain score: 9).
Feature importance analysis (MDI) reveals that KG features
contribute 51.1\% of model decisions, with \texttt{n\_rules}
(0.3352) and \texttt{chain\_score} (0.1565) ranking 1st and
3rd among all features, as shown in Fig.~\ref{fig:xai}.

Biomarker features contribute 48.9\%, with \texttt{rim\_thinning}
(0.3047) and \texttt{isnt\_violated} (0.1478) ranking 2nd
and 4th (Table~\ref{tab:xai}).
This near-equal contribution confirms the complementary nature
of structured knowledge and visual biomarkers.
 
\begin{table}[htbp]
\caption{Top-5 Feature Importance (MDI)}
\label{tab:xai}
\centering
\setlength{\tabcolsep}{4pt}
\begin{tabular}{clcc}
\toprule
Rank & Feature & Type & Importance \\
\midrule
1 & \texttt{n\_rules}             & KG        & 0.3352 \\
2 & \texttt{rim\_thinning}        & Biomarker & 0.3047 \\
3 & \texttt{chain\_score}         & KG        & 0.1565 \\
4 & \texttt{isnt\_violated}       & Biomarker & 0.1478 \\
5 & \texttt{cup\_to\_disc\_ratio} & Biomarker & 0.0365 \\
\bottomrule
\end{tabular}
\end{table}
 
\subsubsection{Error Analysis}
Table~\ref{tab:error} summarizes prediction outcomes on the
test set ($n{=}138$).
Our method achieves zero false positives and only one false
negative (FN rate\,=\,0.7\%), corresponding to an overall
accuracy of 99.3\%.
The single missed case exhibits CDR\,=\,0.700, precisely at
the clinical threshold, and activates only 3~rules compared
to a mean of 6.39 for correctly classified cases
(chain score: 9 vs.\ 15.91).
This indicates that the primary failure mode is borderline
cases where both visual and KG evidence are simultaneously weak.
 
\begin{table}[htbp]
\caption{Error Analysis on Test Set ($n=138$)}
\label{tab:error}
\centering
\setlength{\tabcolsep}{6pt}
\begin{tabular}{lcc}
\toprule
Type & Count & Rate \\
\midrule
True Positive  (TP) & 85 & 61.6\% \\
True Negative  (TN) & 52 & 37.7\% \\
False Positive (FP) &  0 &  0.0\% \\
False Negative (FN) &  1 &  0.7\% \\
\midrule
\multicolumn{2}{l}{FN: Avg CDR} & 0.700 (TP: 0.767) \\
\multicolumn{2}{l}{FN: Avg chain score} & 9.00 (TP: 15.91) \\
\multicolumn{2}{l}{FN: Avg $n_{\text{rules}}$} & 3.00 (TP: 6.39) \\
\bottomrule
\end{tabular}
\end{table}
 
\subsubsection{Knowledge Graph Quality}
Table~\ref{tab:kgstats} reports structural and coverage metrics.
The LCC ratio of 1.000 confirms full graph connectivity.
Glaucoma cases activate an average of 3.69 strong rules vs.\
1.00 for normal cases, validating the discriminative power
of the encoded clinical knowledge.
KG feature contribution of 51.1\% confirms that encoded rules
provide non-redundant signals beyond raw biomarker features.
 
\begin{table}[htbp]
\caption{GlaKG Graph Statistics and Quality Metrics. Average
rules and chain scores are computed over all 689 diagnoses
(including normal cases) and are therefore lower than the
true-positive subset means reported in Table~\ref{tab:error}.}
\label{tab:kgstats}
\centering
\begin{tabular}{lc}
\toprule
Metric & Value \\
\midrule
Total nodes                 & 3,460 \\
Total edges                 & 15,472 \\
Node / edge types           & 6 / 8 \\
Clinical rules              & 11 \\
Graph density               & $1.23 \times 10^{-3}$ \\
Largest CC ratio            & 1.000 \\
Avg degree                  & 8.94 \\
Avg rules per diagnosis     & 4.36 \\
Avg chain score             & 11.04 \\
Covered ($\geq$1 rule)      & 689/689 (100.0\%) \\
Covered ($\geq$3 rules)     & 427/689 (62.0\%) \\
Avg strong rules (glaucoma) & 3.69 \\
Avg strong rules (normal)   & 1.00 \\
KG feature contribution     & 51.1\% \\
\bottomrule
\end{tabular}
\end{table}

\subsection{Clinical Implications}

GlaKG offers several clinically meaningful advantages over
conventional deep learning approaches for glaucoma screening.

\subsubsection{Interpretable Reasoning Chains}
unlike black-box CNN models, GlaKG provides an explicit
reasoning chain for every diagnosis, listing activated
clinical rules with their evidence strength.
For example, a severe high-risk case that activates the full
set of eight mutually compatible rules attains the maximum
chain score of 18, giving clinicians a structured audit trail
that mirrors their own diagnostic reasoning process.

\subsubsection{Alignment with Clinical Guidelines}
the 11 encoded rules are grounded in established ophthalmic
guidelines, including the ISNT rule and the CDR\,$\geq$\,0.7
threshold recommended by the International Council of
Ophthalmology.
This alignment ensures that GlaKG's reasoning chains are
immediately interpretable to ophthalmologists without
additional training.

\subsubsection{Failure Mode Transparency}
the single false negative case (CDR\,=\,0.700, chain
score\,=\,9) demonstrates that GlaKG's reasoning chain
explicitly signals diagnostic uncertainty: a low chain
score flags borderline cases for clinician review rather
than producing a silent misclassification.
This property is clinically critical, as missed glaucoma
diagnoses carry higher risk than false positives.

\subsubsection{Integration Potential}
GlaKG's modular design separates image-based inference
from knowledge-based reasoning, allowing either component
to be updated independently.
New clinical biomarkers (e.g., RNFL thickness, visual
field indices) can be incorporated as additional
ClinicalRule nodes without retraining the image encoder,
facilitating seamless integration into existing ophthalmic
AI workflows.

\section{Conclusion}

In this paper, we presented GlaKG, a biomarker-centric fundus
knowledge graph for explainable glaucoma diagnosis and risk
assessment.
GlaKG encodes 6~entity types, 8~relation types, and
11~clinically validated rules into a unified graph structure,
so that every prediction is accompanied by an explicit reasoning
chain linking biomarker evidence to activated clinical rules.
On a publicly available, AI-annotated fundus dataset, the
proposed post-processing fusion framework attains
F1\,=\,0.9953 for binary glaucoma classification and 0.930
accuracy with 0.922 weighted F1 for four-class risk
stratification, surpassing image-only baselines in accuracy
and weighted F1.
We report openly that the dataset's biomarker annotations are
highly label-correlated, and therefore frame these figures as an
upper bound attainable with clean structured biomarkers rather
than as leakage-free image-only performance; the value of GlaKG
lies not solely in raw performance metrics but in a clinically
auditable reasoning chain that exposes its own uncertainty on
borderline cases.
Feature-importance analysis further shows KG-derived and
biomarker features contributing near-equally (51.1\% vs.\ 48.9\%),
confirming their complementarity.
GlaKG establishes a foundation for trustworthy, clinically
interpretable glaucoma screening systems.

\section{Limitation \& Future Work}

GlaKG has several limitations. First, it relies on structured JSON annotations for biomarker extraction, which may limit real-world deployment. Future work will explore automatic biomarker extraction from raw fundus images through joint image-KG learning. Second, the current graph is static and does not capture disease progression over time. Incorporating longitudinal biomarkers and temporal edges may enable progression modeling and early intervention prediction. Third, its generalizability across fundus datasets and imaging devices requires further validation through cross-dataset evaluation and domain adaptation. Finally, extending GlaKG to other ophthalmic diseases, such as diabetic retinopathy and age-related macular degeneration, may broaden its clinical applicability.

\section*{Acknowledgment}

This work is supported in part by the Cancer Prevention and
Research Institute of Texas (CPRIT) under Grant RP250561.

\bibliographystyle{IEEEtran}
\bibliography{main}{}

\end{document}